\documentclass[conference]{IEEEtran}
\IEEEoverridecommandlockouts

\usepackage{cite}
\usepackage{caption}
\usepackage{subcaption}
\usepackage{amsmath,amssymb,amsfonts}
\usepackage{algorithmic}
\usepackage{graphicx}
\usepackage{textcomp}
\usepackage{xcolor}
\usepackage{multicol}
\usepackage{hyperref}
\usepackage{array,multirow,graphicx}

\usepackage{eso-pic}

\def\BibTeX{{\rm B\kern-.05em{\sc i\kern-.025em b}\kern-.08em
    T\kern-.1667em\lower.7ex\hbox{E}\kern-.125emX}}

\begin{document}

\newcommand\AtPageUpperLine[1]{\put(0,\LenToUnit{\dimexpr\paperheight-\topmargin-\headheight-\headsep-#1\relax}){\makebox[\paperwidth]{\centering\textit{20th ACS/IEEE International Conference on Computer Systems and Applications (IEEE AICCSA 2023)}}}}

\AddToShipoutPictureBG*{\AtPageUpperLine{2cm}} 

\title{An Anomaly Behavior Analysis Framework for Securing Autonomous Vehicle Perception\\
\thanks{This work is partly supported by National Science Foundation (NSF) research projects NSF-1624668, (NSF) FAIN-1921485 (Scholarship-for-Service), Department of Energy/National Nuclear Security Administration under Award Number(s) DE-NA0003946, and AGILITY project 4263090 sponsored by KIAT (South Korea).}
}

\author{\IEEEauthorblockN{1\textsuperscript{st} Murad Mehrab Abrar}
\IEEEauthorblockA{\textit{Dept. of Electrical and Computer Engineering} \\
\textit{The University of Arizona}\\
Tucson, Arizona, USA\\
abrar@arizona.edu}
\and
\IEEEauthorblockN{2\textsuperscript{nd} Salim Hariri}
\IEEEauthorblockA{\textit{Dept. of Electrical and Computer Engineering} \\
\textit{The University of Arizona}\\
Tucson, Arizona, USA \\
hariri@ece.arizona.edu}
}


\maketitle

\begin{abstract}
As a rapidly growing cyber-physical platform, Autonomous Vehicles (AVs) are encountering more security challenges as their capabilities continue to expand. In recent years, adversaries are actively targeting the perception sensors of autonomous vehicles with sophisticated attacks that are not easily detected by the vehicles’ control systems. This paper proposes an Anomaly Behavior Analysis framework to detect perception system anomalies and sensor attacks against an autonomous vehicle. The framework relies on temporal features extracted from a physics-based autonomous vehicle behavior model to capture the normal behavior of vehicular perception in autonomous driving. By employing a combination of model-based techniques and machine learning algorithms, the proposed framework distinguishes between normal and abnormal vehicular perception behavior. As part of our experimental evaluation of the framework, a depth camera blinding attack experiment was performed on an autonomous vehicle testbed and an extensive dataset was generated. The effectiveness of the proposed framework has been validated using this real-world data and the dataset has been released for public access. To our knowledge, this dataset is the first of its kind and will serve as a valuable resource for the research community in evaluating their intrusion detection techniques effectively.

\end{abstract}


\section{Introduction}

Autonomous Vehicle (AV) is one of the most emerging technologies that is gradually making a prominent appearance in our society through improving Advanced Driver Assistance Systems (ADAS) and mitigating errors caused by human factors in driving \cite{adas}. Studies show that more than 90\% of accidents are caused by human errors, killing roughly 43 thousand and injuring over 2 million in the United States alone in 2021 \cite{critical reasons, National Highway Traffic Safety Administration}. Autonomous vehicles have the potential to significantly reduce these fatal accidents by eliminating many of the mistakes that human drivers make routinely \cite{Traffic accidents with autonomous vehicles}. However, as autonomous vehicles become increasingly integrated into our daily lives, ensuring their security and resilience against potential cyberattacks is a critical concern.

Autonomous vehicles or self-driving cars are sensor-enriched vehicles capable of sensing the environment and navigating safely with little or no human input by incorporating vehicular automation \cite{Cooperative control of heterogeneous connected vehicle platoons, av implementation predictions}. Vehicular automation is achieved through a combination of five key components: (1) Perception, (2) Localization and Mapping, (3) Path Planning, (4) Decision Making, and (5) Vehicle Control \cite{Autonomous intelligent vehicles}. Among these components, the perception system plays a vital role in autonomous driving by providing essential environmental information for the vehicle to make decisions and navigate safely to the desired destination \cite{perception planning control}. Perception system heavily relies on an array of perception sensors, such as Camera, LiDAR (Light Detection and Ranging), Sonar, Radar, etc. to understand and interpret the surrounding environment, identify objects, and estimate positions. Unfortunately, these perception sensors are vulnerable to cyberattacks that can compromise their accuracy and reliability, posing significant risks to the overall safety and security of autonomous vehicles. Extensive research conducted over the past few years has revealed techniques to attack vehicular perception sensors, including LiDARs \cite{cao}, Cameras \cite{i can see the light}, Ultrasonic Sensors, and Radars \cite{Can you trust autonomous vehicles}. Sensors may deviate from their calibrated positions due to attacks and mishaps. Such attacks and mishaps can cause perception errors that make sensors fail to understand the surrounding driving environment correctly. In addition, adversaries have the capability to perform these attacks stealthily and remotely, which makes them more challenging to detect. 

Being motivated by the challenges, this paper proposes an Anomaly Behavior Analysis framework to detect perception sensor attacks in autonomous vehicles. This framework aims to identify anomalous behavior patterns exhibited by the vehicular perception sensors, indicating potential attacks or compromises in their functioning. By leveraging machine learning algorithms and model-based techniques, the proposed framework can distinguish between normal and abnormal perception behaviors, facilitating the effective detection of anomalies and attacks. The main contributions of this paper are listed as follows:

\begin{itemize}
    
    \item The paper presents an Anomaly Behavior Analysis framework to secure autonomous vehicles against perception anomalies and attacks. The framework uses temporal features extracted from a physics-based autonomous vehicle behavior model to capture normal vehicular perception. 
    
    \item The paper offers ``AVP-Dataset: Autonomous Vehicle Perception Attack Dataset", a standard and publicly available research dataset with both normal and abnormal instances of vehicular perception. The dataset is collected from practical field experiments of real-world depth camera attacks performed on a commercially available autonomous vehicle testbed.
    
\end{itemize}

The rest of the paper is organized as follows: Section \ref{sec2}, covers the related work; Section \ref{sec3} presents the Anomaly Behavior Analysis framework for secure perception in autonomous vehicles; Section \ref{sec4} presents the experiments and key findings, and Section \ref{sec5} concludes the paper. 

\section{Related Work}\label{sec2}

\subsection{Attacks against the Perception Sensors}

Perception sensors are highly susceptible to remote attacks without the need for physical access \cite{potential cyberattacks}. Intelligent remote attacks, such as spoofing and signal absorption using smart materials pose significant challenges as they can evade detection systems that fail to analyze the correlation among heterogeneous sensors \cite{seeing believing}. Recent studies have conducted real-world experiments demonstrating the feasibility of remote attacks on autonomous driving sensors. Petit et al. attacked the camera and LiDAR of a target autonomous vehicle by using low-cost and widely available hardware components like laser pointers and cheap Light Emitting Diodes (LEDs) \cite{petit}. Extending this work, Shin et al. presented a spoofing attack that induces illusions in the LiDAR output and causes the illusions to appear closer than the location of the spoofer \cite{illusion and dazzle}. Cao et al. investigated the possibility of deceiving the Baidu Apollo LiDAR perception module, achieving a notable attack success rate of 75\% \cite{cao}. Yan et al. performed a series of attack experiments on the sensor suite of a Tesla Model S, encompassing blinding attacks on cameras and jamming and spoofing attacks on radar and ultrasonic sensors \cite{Can you trust autonomous vehicles, ultrasonic sensor}.

\subsection{Perception Security}

Based on the methods to detect perception attacks proposed in \cite{petit, Can you trust autonomous vehicles, illusion and dazzle, ultrasonic sensor}, and \cite{attack-resilient}, the current approaches for detecting perception sensor attacks on autonomous vehicles can be categorized as follows: (1) incorporating redundancy by adding more sensors, (2) leveraging inter-vehicle communications to compare sensor measurements, and (3) relying on alternative sensors/ sensor fusion to identify and detect attacks \cite{seeing believing}. While utilizing multiple sensors of the same kind to detect attacks may enhance resilience against random attacks, it is ineffective against intentional attacks specific to a particular sensor type where the attacker has prior knowledge of the sensor. Using inter-vehicle communications (Vehicle to Vehicle or V2V network) to compare sensor measurements is also an ineffective strategy since this requires the victim vehicle to be located within other vehicles’ communication range \cite{milaat, oligeri}. Recent studies have explored the fusion of multiple sensor modalities for improved attack detection, for instance, combining LiDAR and camera sensory data \cite{rgbd, multi-view, 3d proposal}. This technique achieves high accuracy in non-adversarial settings, yet falls short in effectively detecting attacks targeting multiple sensors \cite{tu, camera-lidar}.

Considering the limitations of the existing approaches, there is a need for a practical solution that can comprehensively analyze attacks without relying on specific sensor nodes, redundant sensors, or other vehicles. We claim that the Anomaly Behavior Analysis approach presented in this paper is a more holistic approach that can overcome these limitations and enhance the perception security of autonomous vehicles.

\section{Proposed Framework}\label{sec3}

The proposed Anomaly Behavior Analysis framework is designed to identify perception sensor anomalies and cyberattacks by continuously monitoring the behavior of the vehicle and considering any deviation from the normal behavior as anomalous. The framework relies on a combination of a mathematical model of the vehicle called the \textit{Autonomous Vehicle Behavior Model} and machine learning algorithms to represent the vehicle's normal behavior, as the mathematical model captures the vehicle's state at any instantaneous time $t$. 

\subsection{Anomaly Behavior Analysis Methodology}

\begin{figure}[htbp]
\centerline{\includegraphics[width=7cm]{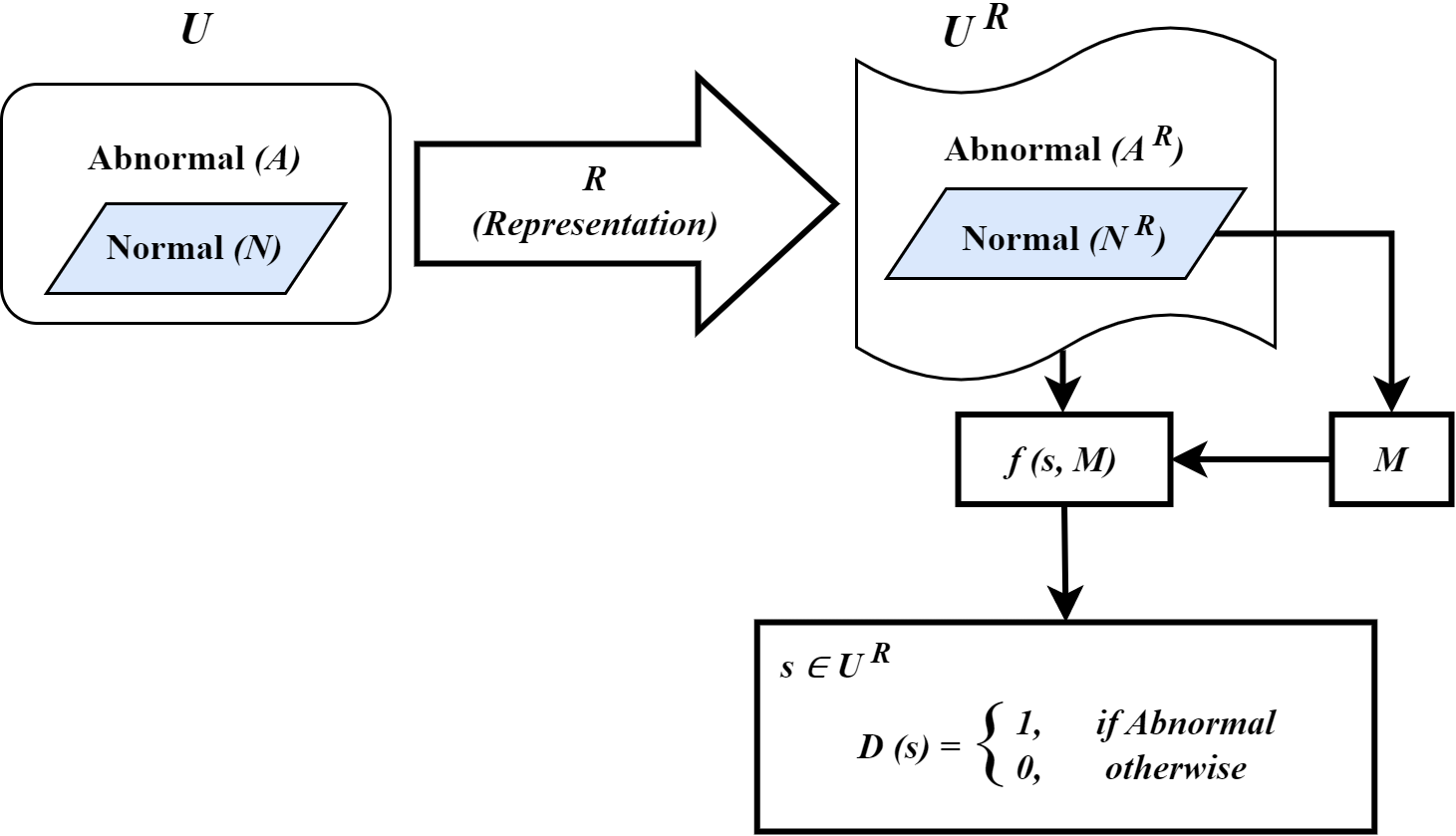}}
\caption{Anomaly Behavior Analysis Methodology} \label{ABAfig}
\end{figure}

Satam et al. \cite{wids} have presented an intrusion detection system for wireless networks based on anomaly detection. We have integrated this system into our Anomaly Behavior Analysis framework to detect a perception sensor attack in autonomous vehicles, which is illustrated in Fig. \ref{ABAfig}. The Anomaly Behavior Analysis approach is defined over a finite set of driving events $U$. Set $U$ is partitioned into two subsets: $Normal$ events $N$ and $Abnormal$ events $A$, such that: $N \cup A = U$ and $N \cap A = \emptyset$. To model $U$, a representation map $R$ is used which maps events in $U$ to patterns in $U^R$. Similarly, $N$ and $A$ sets are mapped to $N^R$ and $A^R$ using the representation map $R$ and can be expressed as: $U \xrightarrow{R} U^R$, 
$N \xrightarrow{R} N^R$, $A \xrightarrow{R} A^R$, and $N^R \cup A^R = U^R$. A detector $D$ is defined as $D = (f, M)$; where $f$ is the normal behavior characterization function expressed as $f: U^R \times M \Rightarrow [0, 1]$, and $M$ is the system memory that stores the normal behavior model extracted from the set of normal events $N^R$. Function $f$ specifies the degree of abnormality of a sample $s \in U^R$ by comparing it with $M$. The higher the value of $f(s, M)$, the more abnormal the sample is. If the value of $f(s, M)$ exceeds a predefined threshold $\mathbb{T}$, detector $D$ raises an alarm indicating the occurrence of an abnormal event. Detector $D$ can be expressed as follows:  
\begin{equation*}
D(s) = 
\begin{cases}
    Abnormal \hspace{1cm} if \quad f(s, M) > \mathbb{T}\\
    Normal \hspace{1.3cm} otherwise
\end{cases}
\end{equation*}


Detection takes place when detector $D$ classifies a sample as abnormal, regardless of whether it is genuinely an anomaly or a regular sample that has been wrongly classified as one. The detector considers two kinds of errors: \textit{False Positives} and \textit{False Negatives}. A \textit{False Positive} detection occurs when a normal sample $s \in N^R$ is detected as abnormal, while a \textit{False Negative} detection occurs when the detector classifies an abnormal sample $s \in A^R$ as normal. 
Our goal is to tune the predefined threshold $\mathbb{T}$ so that the error is minimized. 

\subsection{Autonomous Vehicle Behavior Model}\label{avbm}

The Autonomous Vehicle Behavior Model is a simplified mathematical representation of an autonomous vehicle's dynamics. It gives us insight into the parameters that characterize the normal behavior of the vehicle. A dynamic bicycle model is adopted to explain the dynamics and motion of an autonomous vehicle \cite{vehicle dynamics and control}.

\subsubsection{Dynamic Bicycle Model}\label{dynamicbicyclemodel}

In a 2-dimensional inertial frame, the inertial position coordinates and heading angle of a dynamic bicycle model are defined as \cite{kong}:

\begin{equation}\label{eq1}
v_x  = \dot{x} = v \cos(\psi + \beta)
\end{equation}
\begin{equation}\label{eq2}
v_y = \dot{y} = v \sin(\psi + \beta)
\end{equation}
\begin{equation}\label{eq3}
r = \dot \psi = \frac{v} {l_r} \sin{\beta}
\end{equation}

\begin{figure}
\centerline{\includegraphics[width=8cm]{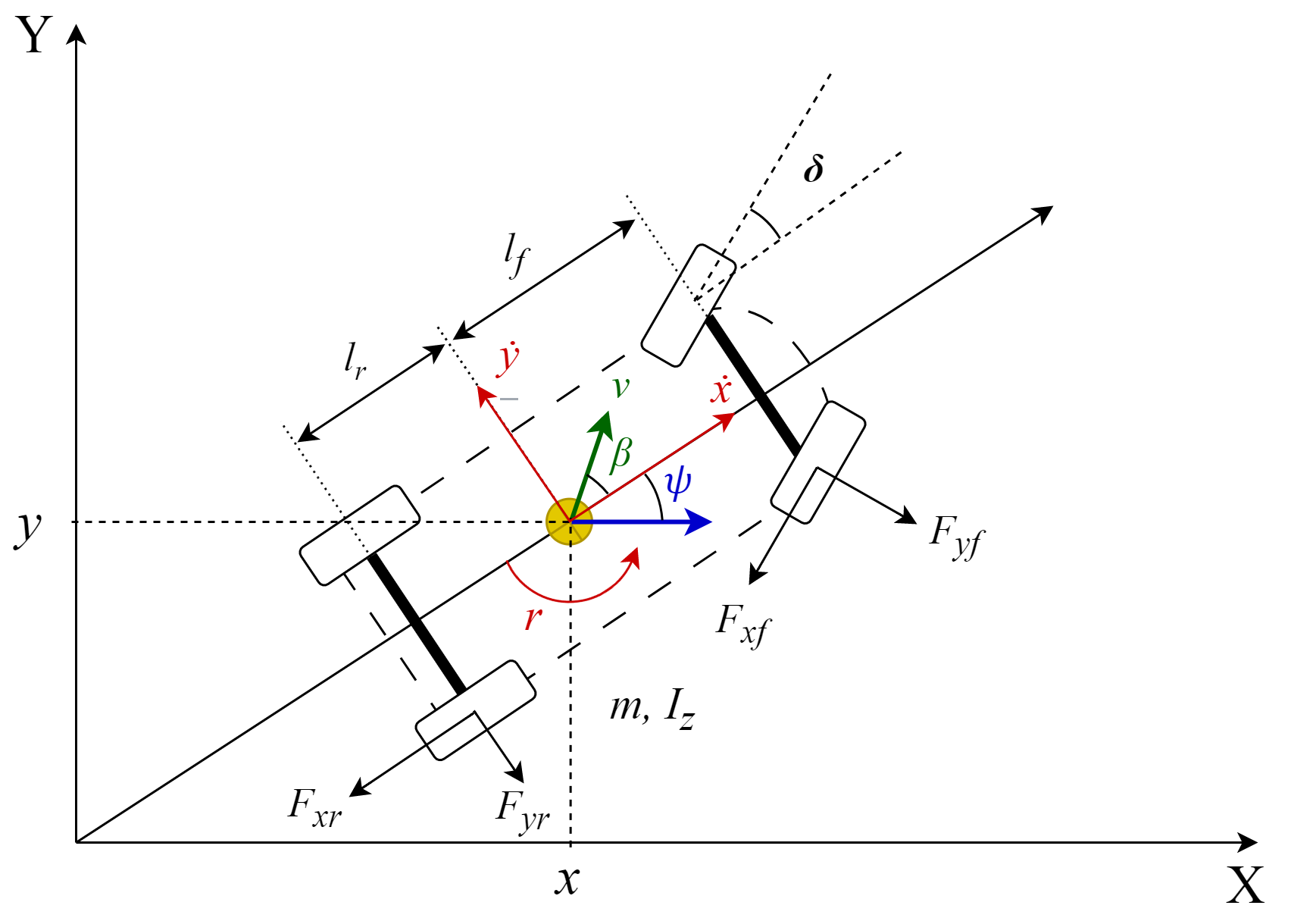}}
\caption{Dynamic Bicycle Model of a vehicle in a 2-dimensional inertial frame}
\label{bicycle}
\end{figure}

Here, \(x\) and \(y\) are the coordinates of the center of mass of the vehicle in a frame (X, Y); \(v_x\) and \(v_y\) are the longitudinal and lateral velocities of the vehicle, respectively; \(\psi\) is the yaw angle or the orientation of the vehicle with respect to the x-axis; \(r\) is the yaw rate; \(\beta\) is the angle of the current velocity of the center of mass with respect to the longitudinal axis of the vehicle; \(\delta\) is the steering angle of the front wheels; and \(l_f\) and \(l_r\) are the distances from the center of mass to the front wheel axle and the rear wheel axle, respectively. The differential equations associated with the dynamic bicycle model of the vehicle are \cite{lateral dynamics, kong}:

\begin{equation}\label{eq4}
\ddot{x} = \dot{\psi} \dot{y} + a_x
\end{equation}
\begin{equation}\label{eq5}
\ddot{y} = - \dot{\psi} \dot{x} + \frac{2}{m} (F_{yf} \cos{\delta} + F_{yr})
\end{equation}
\begin{equation}\label{eq6}
\dot{r} = {\ddot{\psi}} = \frac{2}{I_z} (l_f F_{yf} - l_r F_{yr})
\end{equation}
\begin{equation}\label{eq7}
\dot{X} = \dot{x} \cos{\psi} - \dot{y} \sin{\psi}
\end{equation}
\begin{equation}\label{eq8}
\dot{Y} = \dot{x} \sin{\psi} - \dot{y} \cos{\psi}
\end{equation}

Here, \(\dot{x}\) and \(\dot{y}\) are the longitudinal and lateral velocities of the body frame of the vehicle, respectively; \(a_x\) is the acceleration of the center of the mass of the vehicle; \(\dot{\psi}\) or \(r\) is the yaw rate; \(m\) and \(I_z\) denote the vehicle's mass and yaw inertia, respectively; and \(F_{yf}\) and \(F_{yr}\) denote the lateral tire forces at the front and rear wheels of the vehicle, respectively. From the dynamic bicycle model, Newton-Euler's equations of motion are defined as follows: 
\begin{equation}\label{eq9}
\begin{bmatrix}
\boldsymbol{F}_x \\ 
\boldsymbol{F}_y
\end{bmatrix} 
= m
\begin{bmatrix}
\dot{v_x} - \dot{\psi} v_y \\
\dot{v_y} - \dot{\psi} v_x
\end{bmatrix}
= 
\begin{bmatrix}
-F_{xf} \cos{\delta} - F_{yf} \sin{\delta} - F_{xr} \\
F_{yf} \cos{\delta} - F_{xf} \sin{\delta} + F_{yr}
\end{bmatrix}
\end{equation}
\begin{equation}\label{eq10}
\boldsymbol{\tau} = I_z \ddot{\psi} = I_z \dot{r} = l_f (F_{yf} \cos{\delta} - F_{xf} \sin{\delta}) - l_r F_{yr}
\end{equation}

The dynamics can be simplified by disregarding the aerodynamic resistance acting on the vehicle, which means that the longitudinal tire forces, \(F_{xf}\) and \(F_{xr}\), are zero. Considering a linear tire model, the lateral forces, \(F_{yf}\) and \(F_{yr}\), acting respectively on the front and rear wheels can be defined as:
\noindent
\begin{multicols}{2}
\noindent
\begin{equation}\label{eq11}
F_{yf} = - C_1 \alpha_f 
\end{equation}
\begin{equation}\label{eq12}
F_{yr} = - C_2 \alpha_r
\end{equation}
\end{multicols}

Where \(C_1\) and \(C_2\) are the cornering stiffness coefficients of the front and rear wheels, respectively; and \(\alpha_f\) and \(\alpha_r\) are the slips angles of the front and rear wheels, respectively. Assuming these angles to be small, we obtain:
\noindent
\begin{multicols}{2}
\noindent
\begin{equation}\label{eq13}
\alpha_f = \frac{v_y + l_f r}{v_x} - \delta
\end{equation}
\begin{equation}\label{eq14}
\alpha_r = \frac{v_y - l_r r}{v_x}
\end{equation}
\end{multicols}

Finally, a simplified non-linear state space representation of the lateral dynamics of the vehicle can be expressed as follows:
\begin{equation}\label{eq15}
\begin{bmatrix}
\dot{v_y} \\
\dot{r}
\end{bmatrix} 
= 
\begin{bmatrix}
\textbf{A} & \textbf{B} \\
\textbf{C} & \textbf{D}
\end{bmatrix}
\begin{bmatrix}
v_y \\
r
\end{bmatrix}
+
\begin{bmatrix}
\textbf{E} \\
\textbf{F}
\end{bmatrix}
\delta
\end{equation}

\noindent
\begin{multicols}{2}
\noindent
\begin{equation*}
\textbf{A} = \frac{C_{yf} + C_{yr}}{m v_x}
\end{equation*}
\begin{equation*}
\textbf{B} = \frac{l_f C_{yf} - l_r C_{yr}}{m v^2_x}
\end{equation*}
\noindent
\begin{equation*}
\textbf{C} = \frac{l_f C_{yf} - l_r C_{yr}}{I_z}
\end{equation*}
\begin{equation*}
\textbf{D} = \frac{l^2_f C_{yf} - l^2_r C_{yr}}{I_z v_x} 
\end{equation*}
\noindent
\begin{equation*}
\textbf{E} = - \frac{C_{yf}}{m v_x}
\end{equation*}
\begin{equation*}
\textbf{F} = - \frac{l_f C_{yf}}{I_z} 
\end{equation*}
\end{multicols}

\subsubsection{State Estimation}
For ease of description, we can rewrite equation \ref{eq15} in the following matrix form-
\begin{equation}\label{eq16}
    \dot{\textbf{X}}(t) = f_{state}(\textbf{X}(t), u(t))
\end{equation}

Where the current lateral dynamic state $\textbf{X}(t) = [v_y, r]^t$, control input $u(t) = [\delta]^t$, $f_{state}$ is the nonlinear function reproducing equation \ref{eq15}, and $\dot{\textbf{X}}(t)$ is the predicted next state of the vehicle. 
Since the next state can be computed by using the current state parameters and the instantaneous input, the state of the vehicle can be estimated using equation \ref{eq16}. Thus, the normal behavior of the vehicle is characterized by the state parameters, $v_y$ and $r$, and the steering angle $\delta$.

\section{Experiments}\label{sec4}
\begin{figure}[htbp]
\centerline{\includegraphics[width=8.5cm]{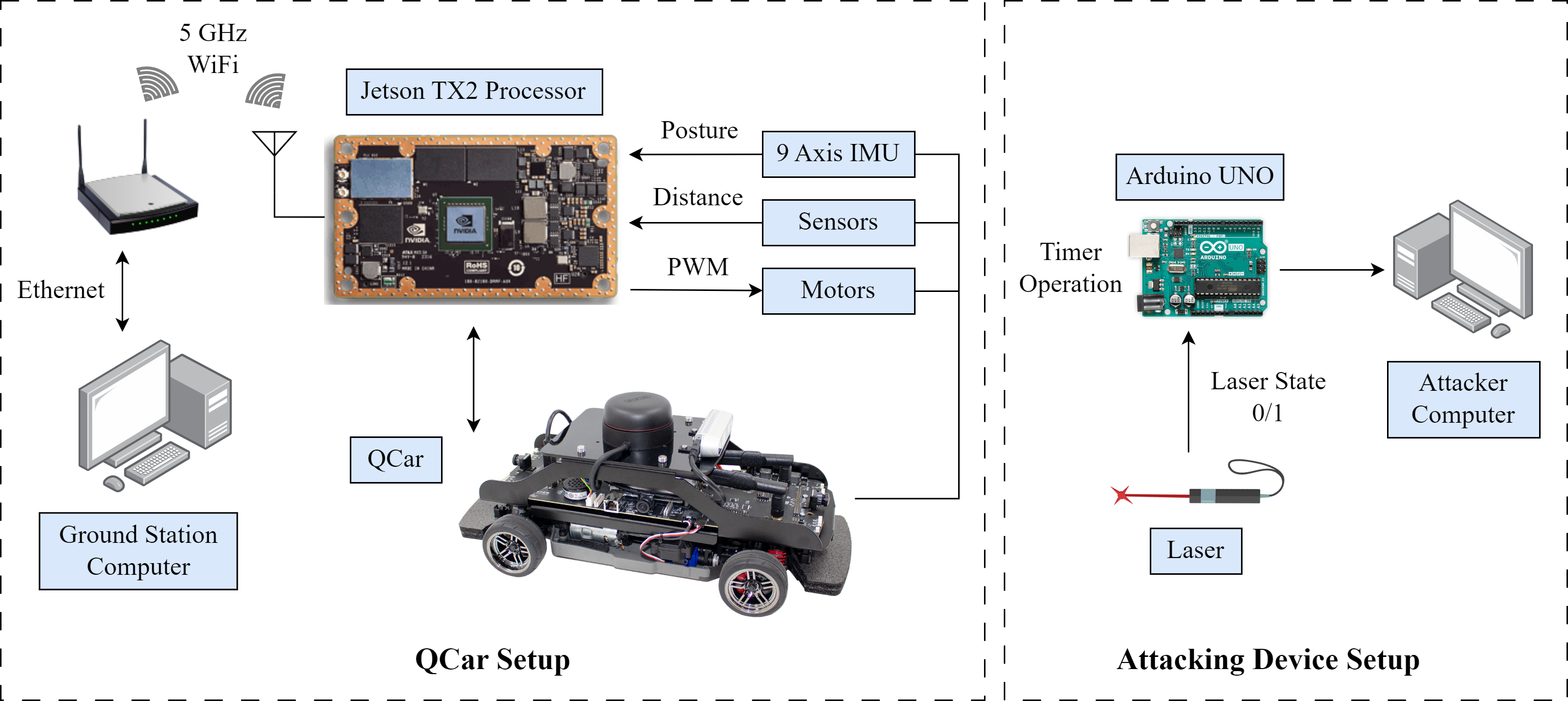}} 
\caption{Architecture of the experimental setup}
\label{qcar setup}
\end{figure}

To validate the proposed Anomaly Behavior Analysis framework for secure perception of autonomous vehicles, experiments have been conducted using the Quanser self-driving car (QCar) robotic platform \cite{qcar}. A real-life depth camera blinding attack was performed on the QCar testbed to collect an extensive dataset for our experimental analysis. The architecture of the experimental setup is shown in Fig. \ref{qcar setup}.

\subsection{QCar Experimental Setup} 
The QCar self-driving robotic testbed incorporates a drive motor and a steering servo motor for its motion. It is equipped with 4 vision cameras, a 2-dimensional LiDAR, and an Intel Realsense depth camera. The pose measurements of the testbed are obtained using an onboard 9-axis Inertial Measurement Unit (IMU). The control algorithm is developed in Simulink, then compiled into C-code and executed on an embedded Linux-based system. The testbed is powered by an onboard NVIDIA Jetson TX2 processor. The processor receives control inputs from the ground station computer via Wi-Fi and transmits the collected IMU and sensor data to the ground station computer.

\subsection{Depth Camera Blinding Attack on QCar} 
The attack was targeted at the Intel Realsense depth camera of the QCar to hamper its braking capabilities at obstacles during autonomous driving \cite{intel realsense}. This attack employed a laser blinding technique as explained by Yan et al. \cite{Can you trust autonomous vehicles}. The experimental setup of the attacking device consisted of a wide-beam laser accompanied by a timer that recorded the timestamps of the laser's operation. The attacking device generated data corresponding to the laser's state, with a value of 1 indicating its activation, and 0 indicating its deactivation. To ensure accurate time synchronization, both the QCar and the attacking device maintained an identical data populating rate and were activated simultaneously. However, during the data preprocessing stage, certain entries were removed to eliminate timing errors and maintain the consistency of timestamps between the two setups.

\subsection{AVP-Dataset}

The AVP-Dataset (Autonomous Vehicle Perception Dataset) is prepared by incorporating the parameters obtained from the Autonomous Vehicle Behavior Model, as detailed in Section \ref{avbm}. The Autonomous Vehicle Behavior Model highlights the essential parameters necessary for estimating the present and future states of the vehicle, thereby offering guidance on the specific data features to be collected for our experimental analysis. Subsequently, data collection modules were developed in the QCar software interface that allowed us to collect real-time sensor data from the testbed and prepare the AVP-Dataset. The full AVP-Dataset is publicly available at \cite{avp online}.

The AVP-Dataset is comprised of three subsets: Subset 1 contains normal data points of 45,001 entries, while Subset 2 and Subset 3 contain abnormal or attack data points of 16,119 entries and 27,856 entries respectively. The normal data encompasses various scenarios, including normal autonomous driving at different velocities, obstacle detection and braking, and autonomous obstacle following. The abnormal data encompasses the depth camera attack during autonomous driving and braking at obstacles. The features of the AVP-Dataset are summarized in Table \ref{avp features}.

\begin{table}[h]
\caption{Features of the AVP-Dataset}\label{avp features}
\centering
\begin{tabular}{l|c|c}
\hline
\hline
\textbf{Features} & \textbf{Notation} & \textbf{Unit} \\
\hline
Timestamp & $t$ & seconds \\
Arm/Disarm & - & unitless \\
Desired Speed & $v_{des}$ & meter/second \\
Longitudinal Speed & $v_x$ & meter/second \\
Lateral Speed & $v_y$ & meter/second \\
Measured Speed & $v$ & meter/second \\
Obstacle Distance & $d$ & meters \\
Steering Angle & $\delta$ & radians \\
Yaw Angle & $\psi$ & radians \\
Yaw Rate & $r$ & radian/second \\
Throttle & $T$ & percentage \\
\hline
\hline
\end{tabular}
\end{table}

\subsection{Experimental Analysis}

\subsubsection{Experiment 1: State Estimation using Dynamic Bicycle Model}

This experiment establishes a connection between the Autonomous Vehicle Behavior Model and the QCar autonomous vehicle testbed. Here, we estimated the lateral dynamic states of the QCar testbed by employing the dynamic bicycle model described in Section \ref{dynamicbicyclemodel}. The estimated results are then compared with the actual values obtained from the AVP normal dataset. Since a small-scale testbed is used to represent the real vehicle, two assumptions are made in this regard: (1) the distances of the front and the rear wheel axle from the center of mass are equal, and (2) the cornering stiffness coefficients of the front and rear wheels are equal to 1. Table \ref{physical parameters qcar} enlists the physical parameters of the QCar testbed that were used to calculate the system matrices and construct the state space representation.

\begin{table}[htbp]
\caption{Physical Parameters of QCar}\label{physical parameters qcar}
\centering
\begin{tabular} {m{4cm} | m{2cm}}
\hline
\hline
\textbf{QCar Parameters} & \textbf{Values}\\
\hline
Mass, $m$ & 2.7 $kg$ \\
Length, $l$ & 0.39 $m$ \\
Width, $w$ & 0.21 $m$ \\
Yaw moment of inertia, $I_z$ & 0.0441 $kg-m^2$ \\
Distance between the front wheel axle and Center of Mass, $l_f$ & 0.16 $m$ \\
Distance between the rear wheel axle and Center of Mass, $l_r$ & 0.16 $m$ \\
Nominal velocity, $v$ & 1 $m/s$ \\
\hline
\hline
\end{tabular}
\end{table}

By incorporating these parameter values into Equation \ref{eq15}, the resulting dynamic state space representation for QCar  is expressed as follows:
\noindent
\begin{equation}\label{final dynamics QCar}
\begin{bmatrix}
\dot{v_y} \\
\dot{r}
\end{bmatrix} 
= 
\begin{bmatrix}
0.7407 & 0.0 \\
0.0 & 1.1598
\end{bmatrix}
\begin{bmatrix}
v_y \\
r
\end{bmatrix}
+
\begin{bmatrix}
-0.3703 \\
-3.6244
\end{bmatrix}
\delta
\end{equation}

\begin{figure}[htbp]
     \centering
     \begin{subfigure}{0.24\textwidth}
         \centering
         \includegraphics[width = \textwidth]{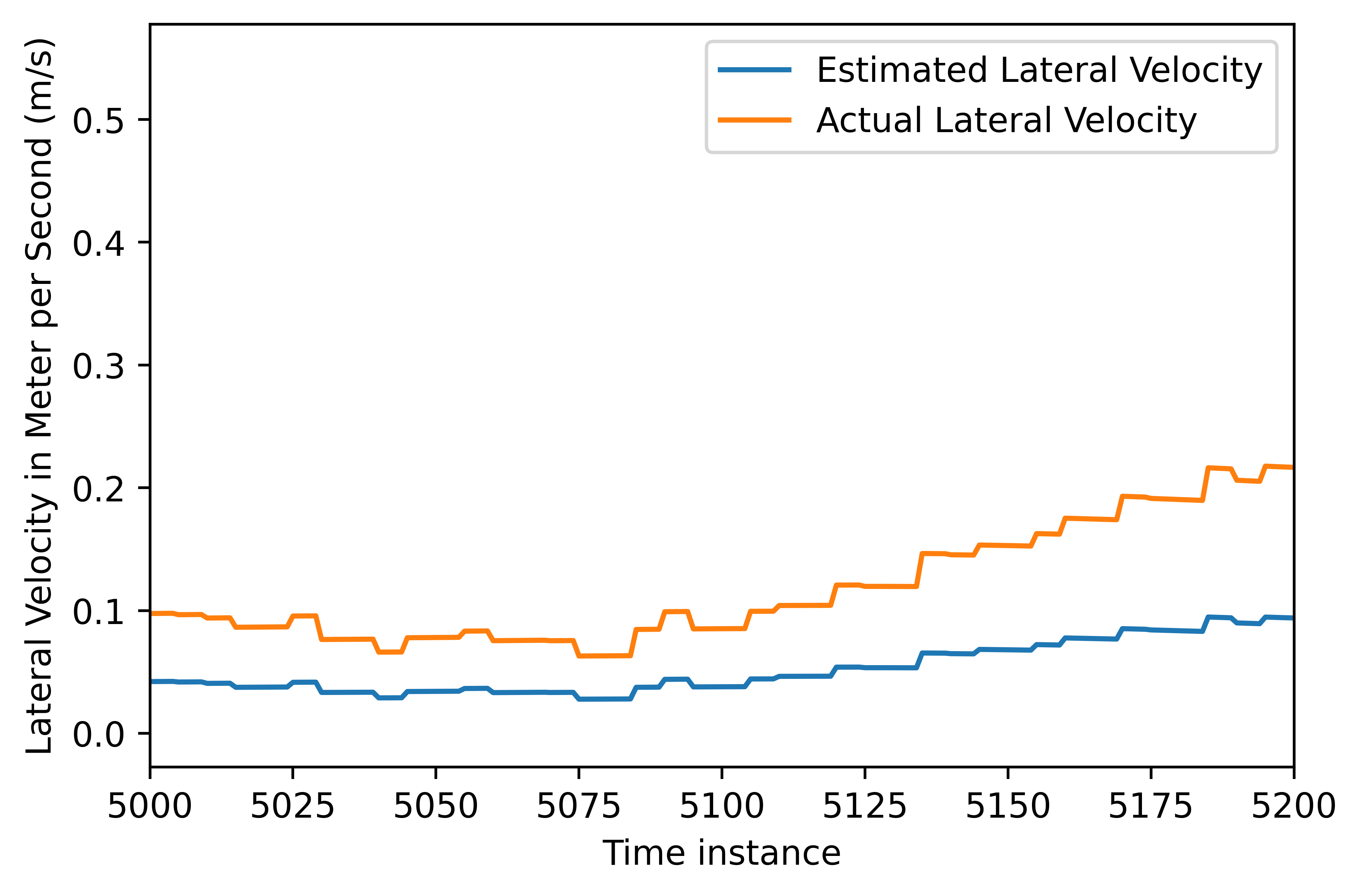}
         \caption{Estimated Lateral Velocity vs Actual Lateral Velocity (QCar)}
         \label{QCar estimated lateral velocity}
     \end{subfigure}
     \hfill
     \begin{subfigure}{0.24\textwidth}
         \centering
         \includegraphics[width = \textwidth]{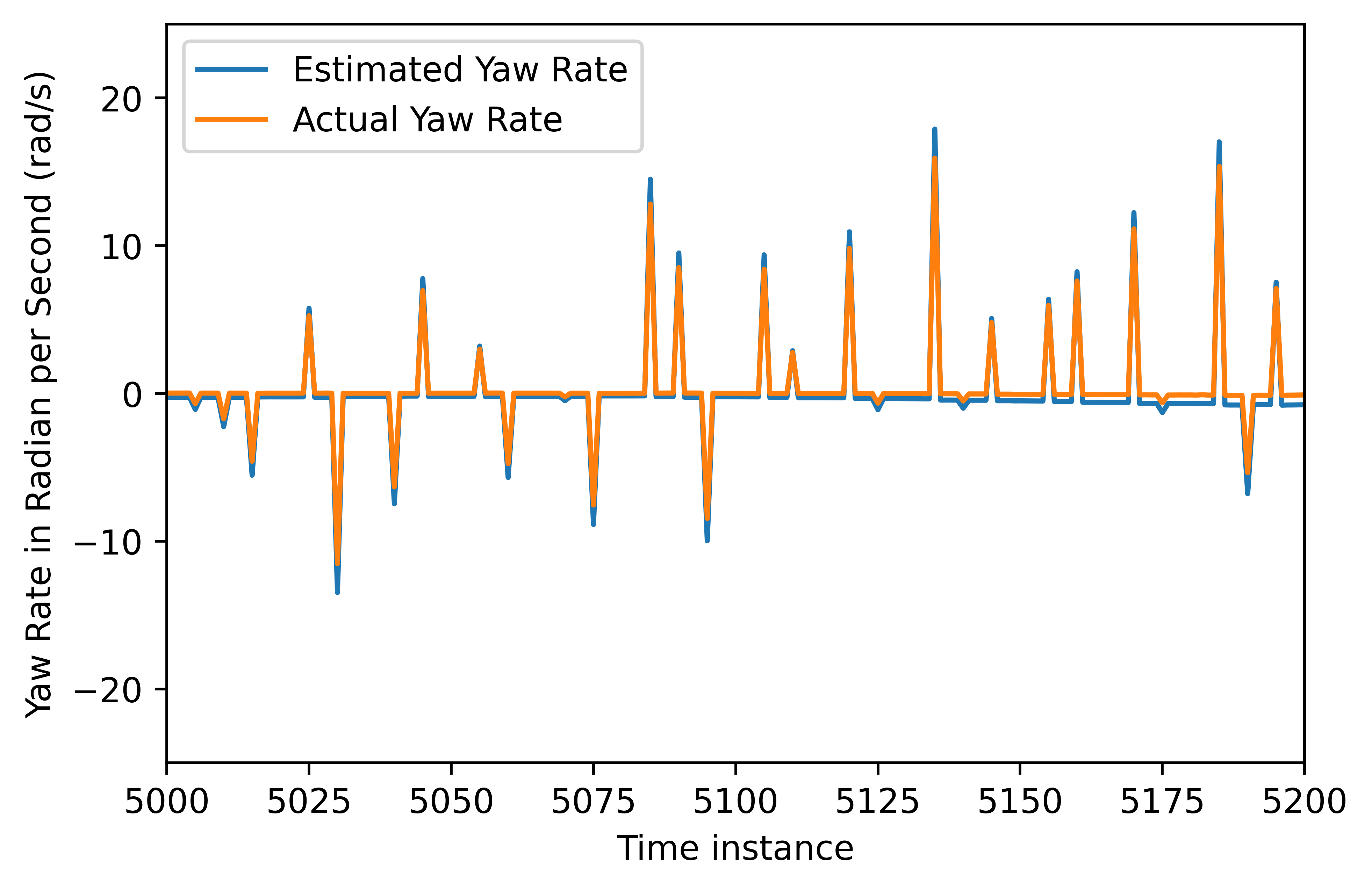}
         \caption{Estimated Yaw Rate vs Actual Yaw Rate (QCar)}
         \label{QCar estimated yaw rate}
     \end{subfigure}
     \caption{Estimated dynamics vs actual dynamics of QCar. Plot limited to instances 5000-5200 for improved visualization}
     \label{QCar Estimation}
\end{figure}

We present a comparative visualization in Fig. \ref{QCar Estimation} depicting the estimated results of the QCar's lateral dynamics along with the corresponding actual values. The results from Fig. \ref{QCar Estimation} demonstrate that the estimated lateral dynamic states obtained using the dynamic bicycle model exhibit a similar distribution to the actual lateral dynamic states of the QCar testbeds. This indicates that the dynamic bicycle model is a suitable representation for capturing the lateral dynamics of the QCar. Consequently, it points out the necessity of adapting the dynamic bicycle model to effectively characterize the behavior of an autonomous vehicle and collect data on the parameters identified by the vehicle behavior model.

Considering that the presented vehicle model accurately estimates the distributions of the next states of the QCar, the state space representation effectively models the QCar testbed. The differences in these values can be accepted based on the assumption that the field experiments added some noise that is not considered in the state equation.

\subsubsection{Experiment 2: Performance Analysis of the Machine Learning Models} 

In this experiment, we evaluated the performance of 7 machine learning binary classification models, namely, Logistic Regression (LR), Random Forest (RF), XGBoost (XGB), K-Nearest Neighbor (KNN), Support Vector Classifier (SVC), Multi-Layer Perceptron (MLP), and Naive Bayes (NB), on the AVP-Dataset. For the experiment, all samples of the dataset (88,976 samples) were used, consisting of 45,001 normal instances (50.57\%) and 43,975 abnormal instances (49.43\%). The performance of the classifiers was evaluated based on 4 metrics: accuracy, precision, recall, and F1 score. Since our objective is to find out if the sample is abnormal, we computed model performances based on the prediction of abnormal data. To ensure a reliable evaluation, stratified 5-fold cross-validation approach was employed to divide the dataset into training and testing sets. The training phase adopted a supervised learning approach, utilizing 80\% of the data (4 folds) for training purposes, while the remaining 20\% (1 fold) was used for testing. Table \ref{mlperformance} summarizes the performance of the classifiers on the AVP-Dataset. The analysis reveals that \textbf{Random Forest} achieved the highest accuracy, precision, recall, and F1 score compared to the other classifiers. Additionally, Logistic Regression, K-Nearest Neighbor, and Multi-Layer Perceptron exhibited comparable F1 scores, following the performance of Random Forest. 

\begin{table}[h]
\caption{Performance of Different Machine Learning Models with Depth Camera Attack on QCar}\label{mlperformance}
\centering
\begin{tabular}{|c|c|c|c|c|c|c|c|}
\hline
\textbf{Metrics} & \textbf{LR} & \textbf{RF} & \textbf{XGB} & \textbf{KNN} & \textbf{SVC} & \textbf{MLP} & \textbf{NB} \\
\hline
Precision & 0.800 & \textbf{0.801} & 0.751 & 0.793 & 0.742 & 0.786 & 0.692 \\
Recall & 0.872 & \textbf{0.894} & 0.876 & 0.877 & 0.913 & 0.874 & 0.923 \\
F1 Score & 0.834 & \textbf{0.845} & 0.809 & 0.833 & 0.819 & 0.828 & 0.791 \\
Accuracy & 0.829 & \textbf{0.838} & 0.795 & 0.826 & 0.800 & 0.820 & 0.759 \\
\hline
\end{tabular}
\end{table}


\subsubsection{Experiment 3: Tuning the Detection Threshold $\mathbb{T}$}

\begin{figure}[b]
\centerline{\includegraphics[width=7.5cm]{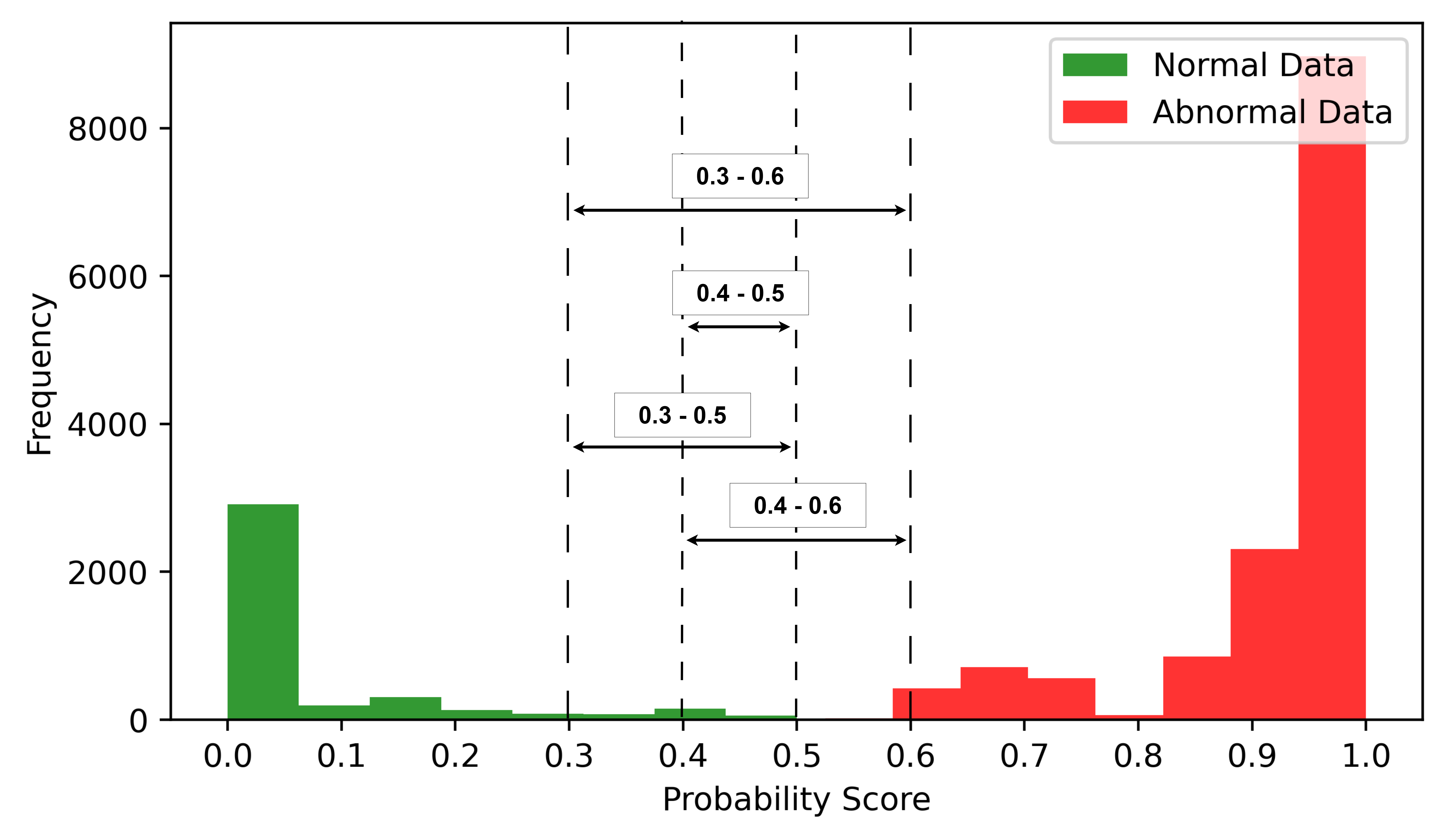}} 
\caption{Comparison of probability score distribution for normal and abnormal data. In 1 fold of test set: Normal data: 3893 instances, abnormal data: 13,902 instances}
\label{margin}
\end{figure}

In this experiment, we conducted an analysis to optimize the predefined threshold $\mathbb{T}$ for the proposed Anomaly Behavior Analysis framework by evaluating the probability scores of both normal and abnormal data. The machine learning model that demonstrated the best performance in Experiment 2, namely \textbf{Random Forest}, was utilized for this purpose. The resulting data were plotted to determine an appropriate margin for placing the detection threshold. Fig. \ref{margin} showcases the probability score distributions of normal and abnormal data overlaid on the same graph. A comparison of these distributions reveals a noticeable separation between the two types of data indicated by the dotted margins.

To further evaluate the performance of different threshold margins, Table \ref{margin table} presents the error rates as the margin is expanded. The findings from Table \ref{margin table} demonstrate that selecting a predefined threshold $\mathbb{T}$ within the range of \textbf{0.4 to 0.5} yields optimal outcomes. Within this range, only 155 instances of normal data are misclassified as abnormal and no attack data are misclassified, resulting in a False Positive Rate of 0.0398 or 3.98\% and a False Negative Rate of 0. Therefore, setting the predefined threshold $\mathbb{T}$ to a value between 0.4 and 0.5 for the Anomaly Behavior Analysis approach ensures an optimal attack detection capability.

\begin{table}[h]
\caption{False Positive (FP) and False Negative (FN) Rates for Different Detection Margins}\label{margin table}
\centering
\begin{tabular}{|c|c|c|c|c|}
\hline
\multirow{2}{*}{\begin{tabular}[c]{@{}c@{}}Detection\\ Margin \end{tabular}} & \multirow{2}{*}{\begin{tabular}[c]{@{}c@{}}Normal Data\\ Misclassified\end{tabular}} & \multirow{2}{*}{\begin{tabular}[c]{@{}c@{}}Attack Data\\ Misclassified\end{tabular}} & \multirow{2}{*}{\begin{tabular}[c]{@{}c@{}}FP \\ Rate\end{tabular}} & \multirow{2}{*}{\begin{tabular}[c]{@{}c@{}}FN\\ Rate\end{tabular}} \\
 &  &  &  &  \\
\hline
\textbf{0.4 - 0.5} & \textbf{155} & \textbf{0} & \textbf{0.0398} & \textbf{0} \\
0.3 - 0.5 & 293 & 0 & 0.0752 & 0 \\
0.4 - 0.6 & 155 & 26 & 0.0398 & 0.00187 \\
0.3 - 0.6 & 293 & 26 & 0.0752 & 0.00187 \\
\hline
\end{tabular}
\end{table}

\section{Conclusion}\label{sec5}

This paper presents an Anomaly Behavior Analysis framework to detect cyberattacks and sensor anomalies in the perception system of autonomous vehicles. The proposed framework combines a model-based approach with supervised machine learning techniques to accurately represent the normal behavior of an autonomous vehicle system and identify any anomalous behavior caused by the perception attacks. The framework is validated on a real-world dataset that we collected from practical depth camera blinding attack experiments on QCar autonomous vehicle testbed. Through experimental analysis, we demonstrated the feasibility of our framework.




\vspace{12pt}

\end{document}